\documentclass{article}
\usepackage{microtype}
\usepackage{graphicx}
\usepackage{subcaption}
\usepackage{booktabs}
\usepackage{hyperref}

\usepackage{tabularx} 

\usepackage[preprint]{icml2026}

\usepackage{amsmath}
\usepackage{amssymb}
\usepackage{mathtools}
\usepackage{amsthm}
\usepackage[capitalize,noabbrev]{cleveref}
\usepackage{multirow} 
\usepackage[table]{xcolor}
\usepackage{xspace} 
\usepackage{booktabs}
\usepackage{pifont}

\newcommand{\sysname}{ARGaze\xspace}

\theoremstyle{plain}

\theoremstyle{definition}

\theoremstyle{remark}

\usepackage[textsize=tiny]{todonotes}

\icmltitlerunning{ARGaze: Autoregressive Transformers for Online Egocentric Gaze Estimation}

\begin{document}

\twocolumn[
  \icmltitle{ARGaze: Autoregressive Transformers for Online Egocentric Gaze Estimation}
  
  \icmlsetsymbol{equal}{*}

  \begin{icmlauthorlist}
    \icmlauthor{Jia Li}{utd}
    \icmlauthor{Wenjie Zhao}{utd}
    \icmlauthor{Shijian Deng}{utd}
    \icmlauthor{Bolin Lai}{gt}
    \icmlauthor{Yuheng Wu}{uwm}
    \icmlauthor{RUijia Chen}{uwm}
    \icmlauthor{Jon E. Froehlich}{uw}
    \icmlauthor{Yuhang Zhao}{uwm}
    \icmlauthor{Yapeng Tian}{utd}
  \end{icmlauthorlist}

  \icmlaffiliation{utd}{Department of Computer Science, University of Texas at Dallas, Richardson, TX, USA.}
  \icmlaffiliation{gt}{College of Computing, Georgia Institute of Technology, Atlanta, GA, USA}
  \icmlaffiliation{uwm}{Department of Computer Science, University of Wisconsin-Madison, Madison, WI, USA}
  \icmlaffiliation{uw}{Allen School of Computer Science, University of Washington, USA}

  \icmlcorrespondingauthor{Jia Li}{Jia.Li@utdallas.edu}
  \icmlcorrespondingauthor{Yapeng Tian}{Yapeng.Tian@utdallas.edu}

  \icmlkeywords{Machine Learning, ICML}

  \vskip 0.3in
]

\printAffiliationsAndNotice{}

\begin{abstract}

Online egocentric gaze estimation predicts where a camera wearer is looking from first-person video using only past and current frames, a task essential for augmented reality and assistive technologies. Unlike third-person gaze estimation, this setting lacks explicit head or eye signals, requiring models to infer current visual attention from sparse, indirect cues such as hand-object interactions and salient scene content. We observe that gaze exhibits strong temporal continuity during goal-directed activities: knowing where a person looked recently provides a powerful prior for predicting where they look next. Inspired by vision-conditioned autoregressive decoding in vision–language models, we propose \sysname, which reformulates gaze estimation as sequential prediction: at each timestep, a transformer decoder predicts current gaze by conditioning on (i) current visual features and (ii) a fixed-length \emph{Gaze Context Window} of recent gaze target estimates. This design enforces causality and enables bounded-resource streaming inference. We achieve state-of-the-art performance across multiple egocentric benchmarks under online evaluation, with extensive ablations validating that autoregressive modeling with bounded gaze history is critical for robust prediction. We will release our source code and pre-trained models.
\end{abstract}
\section{Introduction}
\label{sec_intro}

\begin{figure}[t] 
    \centering
    \includegraphics[width=1.0\linewidth]{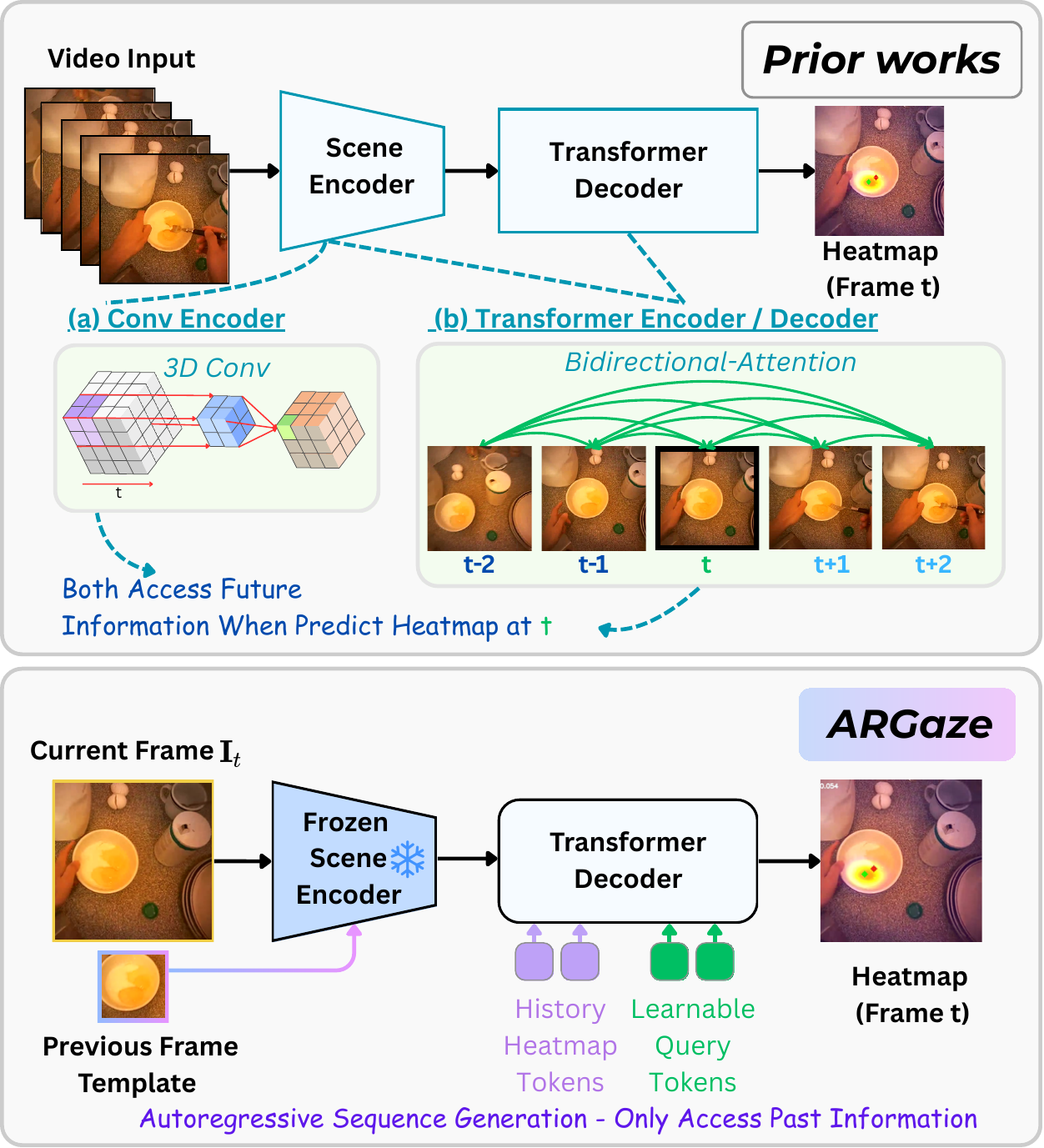}
    \caption{Comparison between offline clip-based and online autoregressive gaze estimation. \textit{Traditional methods (top)} rely on 
    3D convolutions or bidirectional attention, requiring future context that is impractical for real-time deployment. In contrast, \textit{\sysname} reformulates gaze estimation as a causal sequence generation task. By utilizing a 
    visual scene encoder 
    and conditioning on History Heatmap Tokens for temporal stability, \sysname achieves strictly online inference with constant memory overhead, ensuring both responsiveness to new visual observations and fixation stability.}
    \label{fig:Teaser.pdf}
    \vspace{-7mm}
\end{figure}

Egocentric gaze estimation, which involves predicting where a camera wearer is looking from first-person video, is fundamental to human-centric AI systems that understand user intent and provide contextual assistance~\cite{fathi2012learning, li2021eye,grauman2022ego4d, lee2024gazepointar, belardinelli2024gaze, peng2025eye}. Unlike third-person gaze estimation~\cite{recasens2015they,ryan2025gaze,gupta2024mtgs}, which relies on observable head pose and eye cues, egocentric approaches must infer visual attention solely from indirect signals such as scene content, hand–object interactions, and task context~\cite{fathi2012learning, huang2018predicting}, since the wearer’s face is not visible in the camera view. Accurate real-time estimation is critical for applications in augmented reality and assistive robotics, where systems must anticipate the camera wearer's intent and provide proactive guidance as each frame arrives~\cite{damen2022rescaling}.

Recent years have witnessed remarkable progress in egocentric gaze estimation, driven by advances in vision transformers and large-scale egocentric datasets~\cite{grauman2022ego4d, grauman2024egoexo4d}. Current mainstream approaches typically frame egocentic gaze estimation as an offline clip-based spatiotemporal integration problem, employing multi-stage pipelines that  aggregate features across a fixed temporal window to estimate gaze target location~\cite{glc, swingaze}.  However, this offline clip-based paradigm faces limitations in streaming AR and assistive robotics scenarios. Architecturally, most existing models rely on non-causal mechanisms like 3D convolution~\cite{tran20153dconv} and bidirectional self-attention~\cite{vaswani2017attention}, which violate real-time constraints by requiring future temporal context. Their memory requirements also scale with sequence length~\cite{beltagy2020longformer}, making them impractical for processing streaming videos.

Moreover, beyond these architectural constraints, there remains an misalignment between current modeling paradigms and the physiological nature of human gaze. Human vision is characterized by a complex interplay between stable fixations and rapid, discontinuous saccades~\cite{holmqvist2011eye}. While modern architectures effectively aggregate features across temporal windows, they typically treat frames as a parallelized clip of observations rather than a continuous behavioral sequence. These models fail to capture the stability of fixations, causing predictions to fluctuate erratically under visual perturbations (Figure~\ref{fig:reading}).
Addressing this instability requires more than simple temporal smoothing, as a robust system must simultaneously maintain fixation stability and execute precise saccadic jumps when the visual context changes.

Notably, gaze during goal-directed activities exhibits strong temporal continuity, where previous fixations provide a powerful prior for predicting current visual attention. This suggests that gaze prediction should be modeled not as per-frame localization, but as a causal sequential process. These observations raise a central research question: \textit{can egocentric gaze estimation be reformulated as a causal sequence modeling problem that enforces temporal stability, strict online inference, and low computational cost?} This motivates integrating (i) high-level scene understanding to infer candidate gaze targets for both fixations and saccadic shifts with (ii) a structured temporal model that enforces stable, behaviorally grounded gaze dynamics.

To answer this question, we propose \sysname, a novel framework that differs from conventional clip-based pipelines by explicitly modeling gaze as an autoregressive sequence. Instead of treating gaze prediction as independent per-frame localization, \sysname generates gaze heatmaps sequentially, conditioning each prediction on current visual features and a recent history of prior gaze estimates. Visual context is incorporated through cross-attention, analogous to vision-conditioned autoregressive decoding in vision–language models~\cite{li2023blip, alayrac2022flamingo}. This autoregressive formulation allows the model to leverage past fixations as informative priors while remaining responsive to current visual observations.

Crucially, \sysname enforces strict causality and maintains a constant memory footprint via a bounded history window, avoiding computational bottlenecks on long-form videos. Extensive experiments across three benchmarks demonstrate that \sysname achieves a 1.40\% average improvement in F1 score, a 1.82$\times$ inference speedup, and improved robustness by mitigating hand-induced bias and outperforming prior methods under out-of-distribution conditions and real-world testing.
In summary, our contributions are as follows: 1) We introduce \sysname, which reformulates egocentric gaze estimation as a causal autoregressive sequence prediction task, enabling strictly online inference without future context; 2) Our method ensures constant memory usage via a bounded gaze history, supporting efficient long-horizon streaming; 3) \sysname outperforms prior methods on three benchmarks with higher accuracy, faster inference, and stronger robustness, including improved out-of-distribution performance.

\section{Related Work}
\label{sec_related}

\begin{figure*}[t] 
    \centering
    \includegraphics[width=1.0\linewidth]{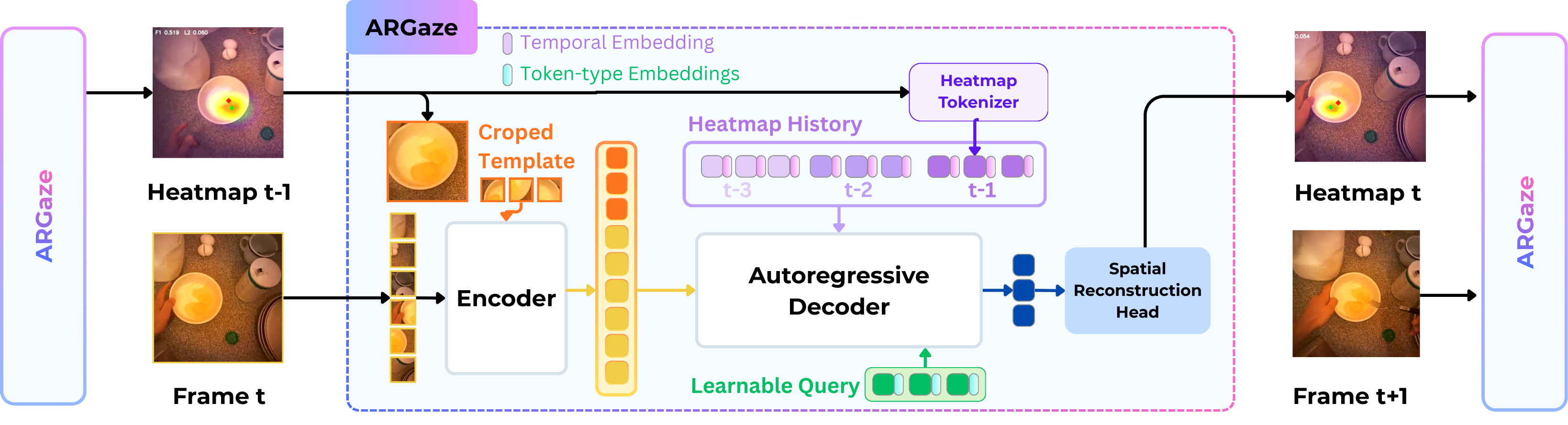}
    \caption{\textbf{Overview of the \sysname framework for online egocentric gaze estimation.} Our model treats gaze prediction as an autoregressive sequence generation task. At each timestep $t$, the Scene Encoder extracts features from the current frame $\mathbf{I}_t$, while the Tracking-aware Template Module provides localized visual priors from previous fixations. The Autoregressive Heatmap Tokenizer converts historical heatmaps $\mathbf{H}_{t-k:t-1}$ into temporal tokens. These inputs are integrated by a Transformer Decoder and mapped back to the spatial domain via a Reconstruction Head to produce the final gaze target heatmap $\mathbf{H}_t$.}
    \label{fig:ARgaze_pipeline}
\end{figure*}

\subsection{Egocentric Gaze Estimation} \label{subsec_egogaze}


Early approaches exploited hand-eye coordination priors, moling gaze as a latent variable conditioned on hand location and head motion~\cite{li2013}. While effective for manipulation tasks, these methods fail in hands-free scenarios such as social interactions where geometric signals are sparse or absent. The shift to deep learning enabled data-driven temporal modeling. Huang et al.~\cite{huang2018} combined spatial saliency streams with LSTM-based temporal modeling to learn task-dependent attention transitions, capturing sequential patterns such as looking at a cup after grasping a bottle. The LSTM encodes gaze history implicitly in its hidden state, eliminating the need for hand-crafted priors. However, recurrent architectures have limited capacity for long-range dependencies compared to attention mechanisms, and subsequent spatiotemporal CNNs~\cite{i3d} struggled with global scene understanding due to local receptive fields~\cite{glc}.
Vision Transformers address these locality constraints through self-attention over spatial patches. Lai et al.~\cite{glc} observed that standard patch-based attention treats all regions equally, missing holistic context needed to disambiguate similar objects. Their Global-Local Correlation module introduces a global scene token that aggregates context and explicitly models its correlation with local patches, substantially improving localization. SwinGaze~\cite{li2023swingaze} extends this with hierarchical transformers for multi-scale objects. While these models can be adapted for streaming inference with causal masking, their architectural design reflects offline training on fixed-length clips with bidirectional temporal context. 
Moreover, standard evaluation settings emphasize clip-level accuracy rather than streaming efficiency or memory scalability.


A key observation motivates our work: gaze exhibits strong temporal continuity during goal-directed activities, where knowing recent fixations provides powerful priors for predicting the next location. However, existing methods either encode this dependency implicitly through recurrent states~\cite{huang2018} or rely on bidirectional attention that violates streaming causality~\cite{glc}. \textit{This exposes a critical gap: how to maintain explicit temporal modeling with strict causal constraints while achieving constant memory footprint for first-person streaming videos.} To bridge the gap, we revisit autoregressive modeling, a paradigm traditionally used in generative tasks for online egocentric gaze estimation.

\subsection{Autoregressive Models in Vision}
\label{subsec_ar_vision}
Autoregressive modeling factorizes sequences as $p(\mathbf{x}_1, \ldots, \mathbf{x}_T) = \prod_{t=1}^T p(\mathbf{x}_t | \mathbf{x}_{<t})$, achieving success in vision primarily for content generation. PixelRNN and PixelCNN~\cite{pixelrnn,pixelcnn} model pixel-level dependencies, VQVAE~\cite{van2017vqvae} introduces discrete tokenization, and recent video models like VideoGPT~\cite{videogpt} and NOVA~\cite{deng2024nova} autoregressively predict future frames. These works optimize for perceptual quality, generating diverse realistic samples.
We introduce a paradigm shift: applying autoregressive modeling to estimation rather than generation. While generative models synthesize visual content, we generate spatial probability distributions for localization. We model $p(g_t | g_{<t}, I_{\leq t})$ where $g_t$ is a gaze heatmap conditioned on visual context $I_t$, optimizing for accurate localization rather than perceptual fidelity. Recurrent architectures like LSTM~\cite{huang2018} also model sequences autoregressively but encode history implicitly in hidden states. Our explicit token-based formulation provides more flexible temporal reasoning through self-attention while maintaining causality. For streaming applications, we introduce bounded context with fixed window size $K$, exploiting the observation that gaze exhibits localized temporal continuity where distant history provides negligible predictive information.

\section{Gaze Estimation as Sequence Generation}
\label{sec:method}

We address \emph{online} egocentric gaze estimation as an autoregressive sequence generation task. Given a continuous stream of egocentric frames, our goal is to predict the gaze heatmap $\mathbf{H}_t$ at each timestep $t$ by modeling the following conditional probability:
\begin{equation}
p(\mathbf{H}_t \mid \mathbf{I}_{t}, \mathbf{H}_{t-k:t-1}),
\label{eq:online_formulation}
\end{equation}
where $\mathbf{I}_t$ represents the current visual frame, and $\mathbf{H}_{t-k:t-1}$ denotes a bounded window of previously generated gaze heatmaps that capture short-term temporal continuity. Unlike standard recurrent models that rely on implicit latent states, or offline models that condition on the entire video sequence $\mathbf{I}_{1:T}$, our formulation treats gaze as an explicit sequence where each prediction is directly conditioned on its own history.
We term this framework as \sysname, which consists of the following main components.

\begin{itemize}
    \item \textbf{Scene Encoder:} Utilizes a pre-trained backbone to extract hierarchical spatiotemporal features from the current egocentric frame and historical visual templates.
    \item \textbf{Autoregressive Heatmap Tokenizer:} Transforms the bounded window of recent gaze history $\mathbf{H}_{t-k:t-1}$ into high-dimensional tokens, explicitly representing temporal fixations as an input sequence for the decoder.
    \item \textbf{Tracking-aware Template Module:} Crops and encodes historical frames around previous gaze locations to provide local visual context, serving as a tracking-style cue to disambiguate current gaze targets.
    \item \textbf{Decoder:} Integrates historical gaze tokens and visual features through a series of self-attention and cross-attention layers, enforcing strict causality while maintaining a constant memory footprint.
    \item \textbf{Spatial Reconstruction Head:} A convolutional upsampling branch that maps the generated sequence tokens back into the spatial domain to produce the final 2D gaze target heatmap $\mathbf{H}_t$.
\end{itemize}

\subsection{Scene Encoder}
\label{subsec:encoder}
\sysname leverages strong visual features from a pre-trained vision foundation model: DINOv3~\cite{simeoni2025dinov3}. The encoding process incorporates both the current observation and historical context to ensure robust performance. From the input frame $\mathbf{I}_t$, we obtain a multiscale feature map that captures high-level semantic context alongside low-level spatial details. As shown in Figure~\ref{fig:ARgaze_pipeline}, this global scene representation is complemented by tracking-aware templates, which consist of localized, high-resolution crops from historical frames centered at previously predicted gaze coordinates. These templates provide specific visual priors that help the model maintain attention on targets amidst rapid camera motion or complex interactions. All extracted features are projected to a latent dimension $d_{model}$ and normalized to yield a unified set of visual tokens, denoted as $\mathcal{F}_t$, which serve as the conditional context for the decoder.

\subsection{Autoregressive Heatmap Tokenizer}
\label{subsec_heatmapTokenizer}
The Autoregressive Heatmap Tokenizer converts the historical gaze sequence into continuous temporal tokens. For each historical heatmap $\mathbf{H}_{t-i} \in \mathbb{R}^{1 \times H \times W}, i \in \{1, \dots, k\}$, we first apply a bilinear downsampling to a resolution of $H_p \times W_p$, matching the spatial grid of visual feature patches. 
To map these single-channel representations into the model's latent space, we employ a pointwise $1\times 1$ convolution that projects the input to the hidden dimension $d_{model}$, followed by Layer Normalization~\cite{ba2016layer}. The resulting feature maps $\mathbf{T}_{t-i} \in \mathbb{R}^{D \times H_p \times W_p}$ are flattened into a sequence of $N = H_p \times W_p$ tokens. 

To enable the model to distinguish between different points in time and functional token roles, we incorporate both \textit{temporal} and \textit{token-type embeddings}. The temporal embeddings explicitly encode the relative chronological order of past fixations, while the token-type embeddings differentiate these historical tokens from the query tokens used for the current prediction. This process transforms the raw gaze history into a structured sequence of tokens that captures the predictive continuity of human attention, which are then concatenated as the input for the causal decoding process.

\subsection{Tracking-aware Template Module}
\label{subsec_template}
The Tracking-aware Template Module exploits the inherent temporal continuity of gaze by providing the decoder with localized visual priors. At each timestep $t$, the module retrieves a bounded history of frames and performs dynamic cropping centered at the previously predicted gaze coordinates. 
To obtain the center of the region of interest at each timestep $t$, we apply a \textit{differentiable soft-argmax} over the predicton heatmap $\mathbf{H}_{t-1}$, yielding normalized coordinates $(x,y)\in[0,1]^2$
This process focuses the model’s attention on a specific region of interest, capturing the localized visual context that is most relevant to the wearer's current focus. 
The extracted crop is resized to the canonical input resolution $H_c \times W_c$ and encoded by the shared frozen backbone to produce a sequence of template tokens. By integrating these high-resolution tracking cues, the model can effectively disambiguate similar-looking objects and maintain stable gaze predictions during periods of rapid ego-motion. This mechanism serves as an explicit spatial bridge between historical fixations and the current visual search, reinforcing the autoregressive consistency of the entire system by allowing the decoder to attend to specific regions identified in the immediate past.

\begin{figure}[t]
    \centering
    \includegraphics[width=0.8\linewidth]{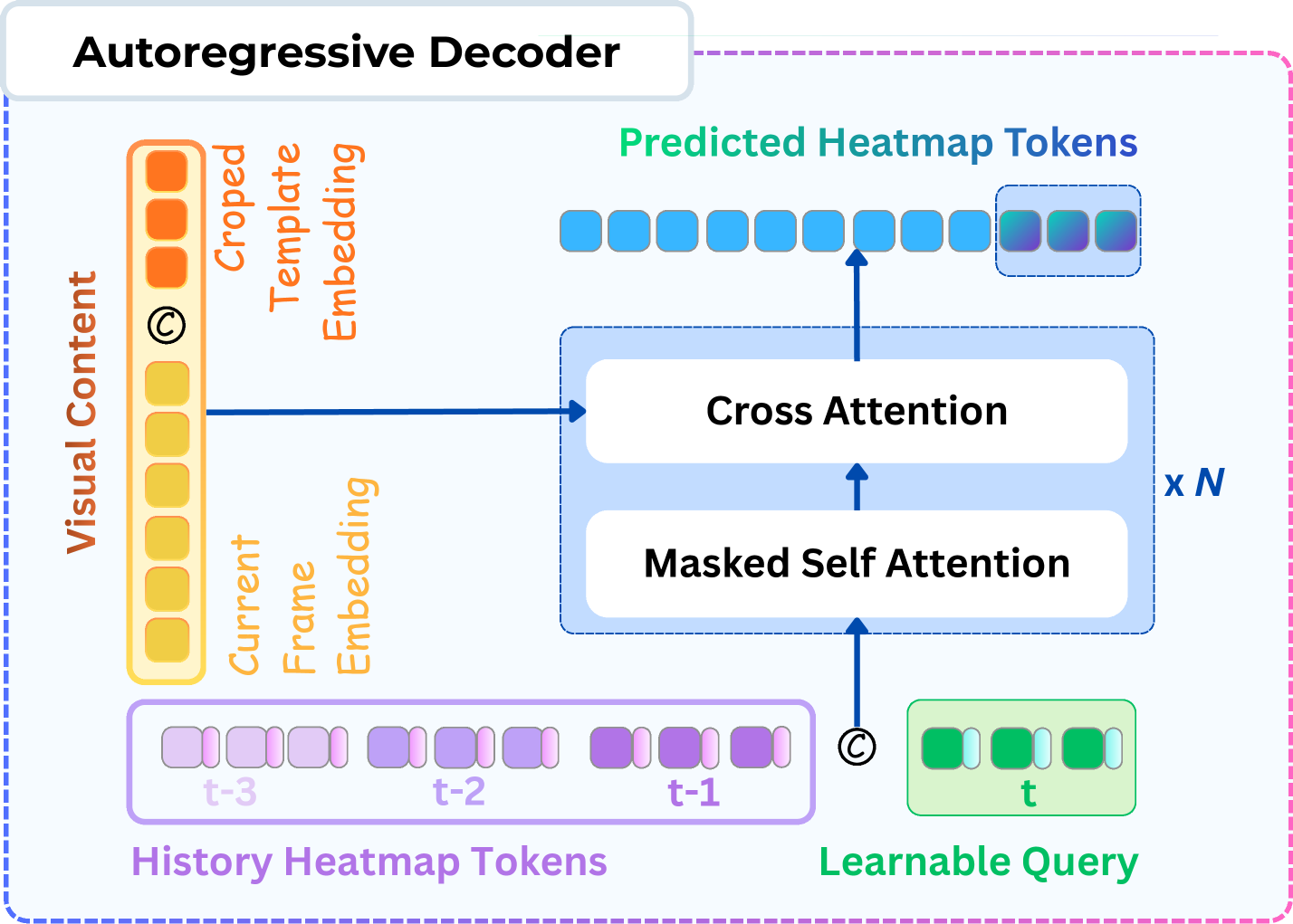}
    \caption{\textbf{Autoregressive Decoder.} The decoder takes $\mathbf{S}_{tgt}=[\mathbf{H}_{t-k:t-1}, \mathbf{q}_t]$ (shown with $k=3$) and applies $N$ masked self-attention and cross-attention layers over visual features (current frame + tracking-aware template embeddings) to predict heatmap tokens with strict causality and constant memory.}

\label{fig:decoder_arch}
\vspace{-7mm}
\end{figure}

\subsection{Decoder}
\label{subsec_decoder}

The Causal Transformer Decoder serves as the central processing unit of \sysname, responsible for integrating historical gaze patterns with current visual features to generate future predictions. As illustrated in Figure~\ref{fig:decoder_arch}, at each timestep $t$, the decoder processes an input sequence $\mathbf{S}_{tgt}$ formed by the concatenation of historical heatmap tokens and a set of learnable query tokens representing the current target frame. 
The learnable query tokens function as informative probes that leverage self-attention layers to aggregate temporal dynamics from the gaze trajectory, thereby maintaining predictive continuity by attending to previous fixations.


To ground these temporal predictions in the current frame, the decoder performs cross-attention over a unified context representation formed by concatenating the spatial tokens  and the localized tracking-aware template tokens from scene encoder. By attending to this unified representation, the decoder enables the query tokens to retrieve global scene context and fine-grained local details simultaneously. 

This decoding process is designed to be causal by operating only on current and past observations, thereby ensuring that the model is suitable for real-time streaming applications. Furthermore, since the decoder only maintains a bounded window of historical tokens, it achieves a constant memory footprint that does not scale with the length of the video stream. This architecture effectively bridges the gap between high-capacity spatial reasoning and the efficiency requirements of unbounded egocentric video processing.

\subsection{Spatial Reconstruction Head}
\label{subsec_head}

The Spatial Reconstruction Head maps the latent representations from the decoder back into the spatial domain to produce the final gaze heatmap $\mathbf{H}_t$. This module operates on the output query tokens, which are first reshaped and permuted to restore the $H_p \times W_p$ spatial grid configuration.

To recover high-resolution spatial details sacrificed during the encoding process, the head employs a series of convolutional upsampling stages. This architecture consists of $3 \times 3$ convolutional layers followed by ReLU activations and transposed convolutions, which progressively increase the feature map resolution. 
Dropout layers are integrated between these stages to mitigate overfitting and promote better generalization, which is crucial for maintaining stable gaze estimation in out-of-distribution (OOD) scenarios.
Finally, a pointwise convolution projects the upsampled features into a single-channel heatmap of resolution $H_h \times W_h$, representing the predicted gaze probability distribution. By directly mapping the generated tokens to the spatial domain, the reconstruction head enables precise attention localization while remaining fully compatible with causal streaming for downstream AR and robotics applications.

\subsection{Training and inference}
\label{subsec_train_inf}

During training, we optimize the model using a sequence-based objective that aligns with its autoregressive nature. The primary objective is to minimize the Kullback-Leibler (KL) divergence between the predicted gaze distribution $\hat{\mathbf{H}}_t$ and the ground-truth heatmap $\mathbf{H}_t$ at each timestep $t$:
$$\mathcal{L}_{kl} = \sum_{t=1}^{T} \sum_{i,j} \mathbf{H}_{t,i,j} \log \left( \frac{\mathbf{H}_{t,i,j}}{\hat{\mathbf{H}}_{t,i,j} + \epsilon} \right),$$

where $i,j$ denote the spatial indices of the heatmap and $\epsilon$ is a small constant added for numerical stability. This pixel-wise distribution matching ensures that the generated heatmaps accurately capture the nature of human visual attention.

To bridge the gap between training with ground-truth history and inference with predicted history, we employ scheduled sampling. During training, the historical tokens $\mathbf{H}_{t-k:t-1}$ are sampled from either the ground-truth heatmaps or the model's generated predictions based on a dynamic probability $p_{ss}$. 
This exposes the decoder to imperfect historical context and encourages it to rely on the current scene tokens to re-align its predictions, yielding stable gaze estimation during rapid saccades in long-term online deployment.

During inference, the model operates in a streaming manner. For each arriving frame $\mathbf{I}_t$, the decoder generates a gaze prediction $\hat{\mathbf{H}}_t$ conditioned on the visual features $\mathcal{F}_t$ and previous estimates. Since the history is maintained within a fixed-size bounded window, the model preserves a constant memory footprint regardless of the total duration of the video stream. This causal loop ensures that the system provides proactive and temporally consistent gaze estimation as the video frame arrives, making it more suitable for real-time integration in mobile and wearable AI devices.

\section{Experiments}

\begin{table*}[ht]
    \centering
    \small
    \setlength{\tabcolsep}{5pt}
    \renewcommand{\arraystretch}{0.9}
    \caption{\textbf{Overall performance under online evaluation.} We report F1, Precision, and Recall (\%) on EGTEA Gaze+, Ego4D, and EgoExo4D. Here, GLC denotes a causal, online adaptation of the original GLC model; implementation details are provided in the appendix. Top-2 results are highlighted.}
    \begin{tabular}{lccc ccc ccc}
        \toprule
        \multirow{2}{*}{\textbf{Method}} &
        \multicolumn{3}{c}{\textbf{EGTEA Gaze+ }} &
        \multicolumn{3}{c}{\textbf{Ego4D }} &
        \multicolumn{3}{c}{\textbf{EgoExo4D (IDD)}} \\
        \cmidrule(lr){2-4}\cmidrule(lr){5-7}\cmidrule(lr){8-10}
        & \textbf{F1} & \textbf{Precision} & \textbf{Recall} &
          \textbf{F1} & \textbf{Precision} & \textbf{Recall} &
          \textbf{F1} & \textbf{Precision} & \textbf{Recall} \\
        \midrule
        Random Prior     & 12.16 &  6.75 & 61.34 &  9.97 &  5.48 & 55.16 & 20.89 & 12.89 & 55.11 \\
        Center Prior     & 21.25 & 13.89 & 45.26 & 26.48 & 17.30 & 56.40 & 46.05 & 36.39 & 62.69 \\
        Dataset Prior    & 26.39& 19.46& 40.99 & 31.90& 24.23& 46.68& 59.84& 64.39& 55.93\\
        \midrule
        AT~\cite{huang2018predicting} & 23.87 & 17.49 & 37.59 & 19.47 & 15.89 & 25.12 & 22.14 & 14.62 & 45.58 \\
        EgoM2P~\cite{li2025egom2p} & 40.14 & \textbf{35.56} & 46.70 & 40.61 & \textbf{35.67} & 47.37 & 67.13 & 67.09 & \textbf{67.18} \\
        GLC~\cite{glc}       & \textbf{41.57} & 32.41 & \textbf{57.94} & \textbf{41.17} & 32.52 & \textbf{56.09} & \textbf{69.94} & \textbf{74.51} & 65.90 \\
        ARGaze(Ours)      & \textbf{44.48} & \textbf{35.22} & \textbf{60.37} &
       \textbf{41.82} & \textbf{33.47} & \textbf{55.71} &
       \textbf{70.58} & \textbf{75.66} & \textbf{66.14} \\
        \bottomrule
    \end{tabular}
    \label{tab:overall_performance}
    \vspace{-15pt}
\end{table*}

\subsection{Experimental Setup}
\textbf{Datasets.}
We used three popular egocentric benchmarks.

\textit{EGTEA Gaze+}~\cite{li2018egtea} contains 28 hours of cooking activities from 32 subjects. We follow the official train/test split, which yields 8,299 training and 2,022 test clips. Videos are provided at 24 fps and a resolution of 640$\times$480. Gaze annotations are provided as 2D screen coordinates, which we normalize to [0,1] and convert to 64$\times$64 spatial heatmaps using Gaussian smoothing. For frames with untracked gaze, we apply a uniform distribution.

\textit{Ego4D (Gaze subset)}~\cite{grauman2022ego4d} is a large-scale egocentric corpus, and we use its gaze-annotated benchmark subset. To construct training clips, we follow prior work~\cite{glc} and apply a temporal sliding window over the video stream. Using a stride of 8 frames yields 183{,}720 training and 15{,}606 test windows. 

\textit{EgoExo4D}~\cite{grauman2024egoexo4d} is a large-scale dataset capturing human daily activities from both egocentric and multiple exocentric perspectives. We utilize the egocentric subset, where 3D gaze vectors provided by the Project Aria Machine Perception Services are projected into 2D screen coordinates and processed into $64{\times}64$ heatmaps following the same protocol as EGTEA Gaze+. To evaluate robustness under distribution shifts, we build an Out-of-Distribution (OOD) benchmark with three controlled axes:
(i) {OOD-Site}: test videos collected at \emph{unseen recording sites};
(ii) {OOD-Task}: test videos from \emph{unseen task categories};
(iii) {OOD-Participant}: test videos performed by \emph{unseen individuals}.
We also report an {IID} test split sampled from the remaining data and designed to follow the same distribution as the training set. In summary, we have 17,397 training samples and four test splits, including IID (4,970), OOD-Site (4,970), OOD-Task (4,970), and OOD-Participant (834).

Beyond standard evaluations, we leverage the extended egocentric streams in Ego4D~\cite{grauman2022ego4d} and EgoExo4D~\cite{grauman2024egoexo4d} to conduct a long-term streaming test, which evaluates the model's inference efficiency over continuous video sequences. The details of this streaming analysis are presented in Table~\ref{tab:efficiency_glc_ours_ego4d}.

\textbf{Baseline.} 
We compare \sysname with static priors, representative baselines AT~\cite{huang2018predicting}, EgoM2P~\cite{li2025egom2p} and the state-of-the-art method GLC~\cite{glc}. For fair comparison, we extend all baselines to online setting. Implementation details are in Appendix~\ref{app:glc_online}. 

\textbf{Evaluation Metircs.}
Following established protocols in egocentric gaze research~\cite{glc}, we adopt the \textit{Adaptive F1 score} as our primary metric to mitigate the saturation issues associated with AUC in long-tailed gaze distributions. We also report precision and recall to provide a comprehensive assessment of the model's spatial localization performance across diverse video sequences.


\subsection{Main Results}
\textbf{Main Results.}
Compared to the causal implementation of GLC \cite{glc}, our framework achieves consistently better overall performance, attaining higher F1 scores across all benchmarks (Table \ref{tab:overall_performance}).
On the EGTEA Gaze+ dataset, \sysname achieves an F1 score of 44.48\%, outperforming the adapted GLC baseline by nearly three percentage points. This advantage persists in the large-scale and unconstrained environments of Ego4D, where our model yields an F1 of 41.82 percent. 
EgoExo4D shows similar gains, with our model reaching 75.6\% precision and 70.58\% F1.
These consistent gains across diverse settings suggest that the autoregressive design is more effective at disambiguating gaze targets in skilled activities where visual cues are complex and head motion is frequent.

\textbf{Efficiency Analysis. }
The comparative results on the Ego4D dataset highlight the significant efficiency gains of \sysname over the baseline GLC model across all performance metrics. All efficiency measurements are conducted on an NVIDIA RTX A6000 GPU. As shown in Table \ref{tab:efficiency_glc_ours_ego4d}, \sysname achieves a substantial 1.82$\times$ speedup in inference throughput, reaching 88.59 FPS, and reduces mean latency from 20.56 ms to 11.29 ms. This performance boost is accompanied by even greater improvements in memory management; \sysname reduces CUDA memory allocation by 3.13$\times$ and reserved memory by 3.81$\times$, requiring only 188.0 MB compared to GLC's 716.0 MB. Furthermore, the reduction in P95 latency (from 25.41 ms to 12.42 ms) suggests that \sysname is not only faster on average but also more consistent and reliable for real-time applications where tail latency is a critical factor.
The high efficiency of \sysname makes it well-suited for resource-constrained environments. Future work will focus on deploying it to edge devices and AR systems.
\begin{table}[t]
\centering
\small
\setlength{\tabcolsep}{4pt}
\caption{Efficiency comparison on long videos.}
\begin{tabular}{lccc}
\hline
 Metric & GLC & \sysname & Improvement \\
\hline
 Inference FPS $\uparrow$        & 48.74 & 88.59 & 1.82$\times$ \\
  Mean Latency (ms) $\downarrow$  & 20.56 & 11.29 & 1.82$\times$ \\
  P95 Latency (ms) $\downarrow$   & 25.41 & 12.42 & 2.05$\times$ \\
  CUDA Alloc (MB) $\downarrow$    & 526.3 & 168.2 & 3.13$\times$ \\
  CUDA Reserved (MB) $\downarrow$ & 716.0 & 188.0 & 3.81$\times$ \\
\hline
\end{tabular}
\label{tab:efficiency_glc_ours_ego4d}
\vspace{-20pt}
\end{table}

\begin{table*}[t]
    \centering
    \small
    \setlength{\tabcolsep}{5pt}
    \renewcommand{\arraystretch}{0.9}
    \caption{\textbf{Out-of-distribution generalization on EgoExo4D under strictly online evaluation.}
    We report F1, Precision, and Recall on the IID split and three OOD splits: participant, site, and task.}
    \label{tab:egoexo4d_ood_generalization}
    \begin{tabular}{lccc ccc ccc ccc}
        \toprule
        \multirow{2}{*}{\textbf{Method}} &
        \multicolumn{3}{c}{\textbf{IID}} &
        \multicolumn{3}{c}{\textbf{OOD Participant}} &
        \multicolumn{3}{c}{\textbf{OOD Site}} &
        \multicolumn{3}{c}{\textbf{OOD Task}} \\
        \cmidrule(lr){2-4}\cmidrule(lr){5-7}\cmidrule(lr){8-10}\cmidrule(lr){11-13}
        & \textbf{F1} $\uparrow$ & \textbf{Prec.} $\uparrow$ & \textbf{Rec.} $\uparrow$ &
          \textbf{F1} $\uparrow$ & \textbf{Prec.} $\uparrow$ & \textbf{Rec.} $\uparrow$ &
          \textbf{F1} $\uparrow$ & \textbf{Prec.} $\uparrow$ & \textbf{Rec.} $\uparrow$ &
          \textbf{F1} $\uparrow$ & \textbf{Prec.} $\uparrow$ & \textbf{Rec.} $\uparrow$ \\
        \midrule
        Random Prior   & 20.89 & 12.89 & 55.11 & 19.91 & 12.15 & 55.09 & 19.97 & 12.19 & 55.26 & 19.77 & 12.08 & 54.46 \\
        Center Prior   & 46.05 & 36.39 & 62.69 & 46.03 & 36.39 & 62.63 & 46.44 & 36.72 & 63.18 & 42.98 & 32.54 & 63.31 \\
        Dataset Prior  & 59.84& 64.39& 55.93& 65.10& 69.98& 60.84& 64.35& 69.18& 60.15& 62.88& 67.61& 58.78\\
        \midrule
        GLC      & 69.94 & 74.51 & 65.90 & 66.23 & 69.10 & \textbf{63.60} & 67.10 & 69.91 & 64.51 & 69.16 & 72.11 & \textbf{66.43} \\
        Ours (AR)      & \textbf{70.58} & \textbf{75.67} & \textbf{66.14} &
                         \textbf{68.22} & \textbf{75.49} & 62.22 &
                         \textbf{69.56} & \textbf{74.76} & \textbf{65.04} &
                         \textbf{71.10} & \textbf{76.62} & {66.31} \\
        \bottomrule
    \end{tabular}
\end{table*}

\begin{figure*}
    \centering
    \vspace{-3mm}
    \includegraphics[width=0.85\linewidth]{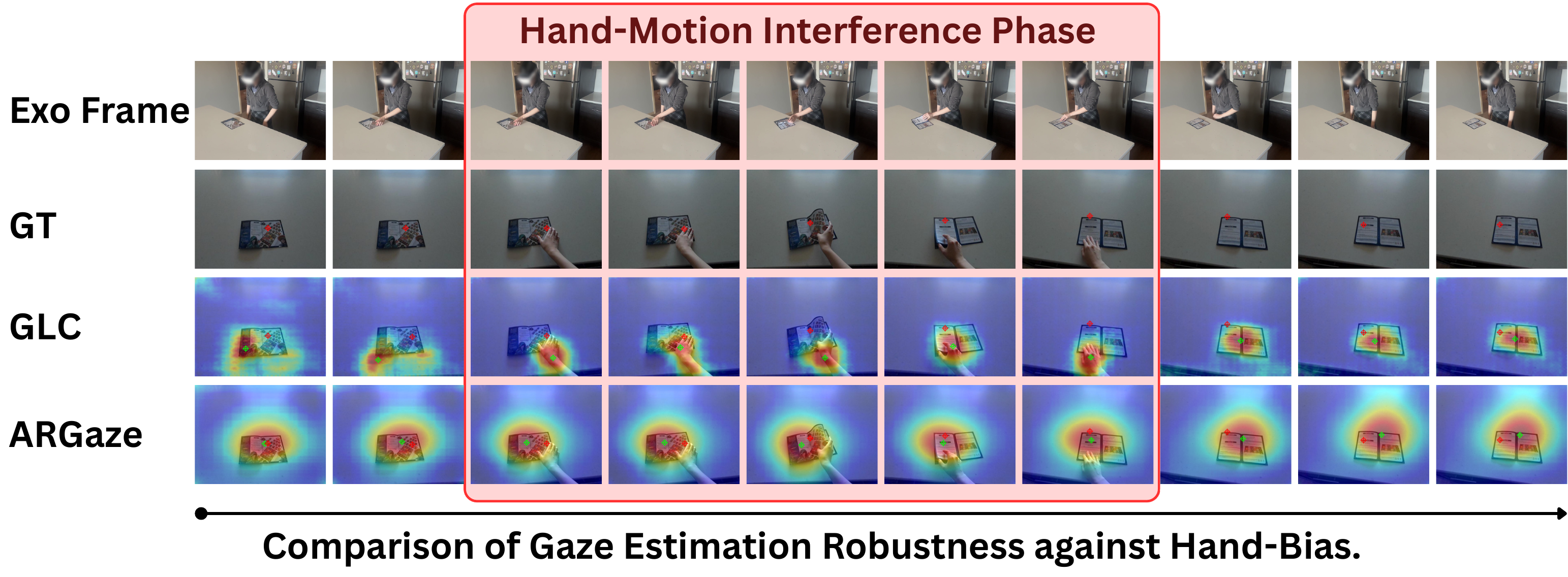}
    \vspace{-5pt}
    \caption{\textbf{\sysname and GLC Regarding Hand-Bias.} We can see that the GLC suffers from significant hand-bias as the predicted gaze erroneously tracks the moving hand. In contrast, \sysname maintains object-centric stability. The red shaded area highlights a period of intensive hand-object interaction where our structured dynamical model ensures a consistent fixation on the task-relevant object. Note that Exo (third-person) frames are not available to the model for egocentric gaze estimation and are shown only for visualization purposes.}
    \label{fig:reading}
    \vspace{-12pt}
\end{figure*}

\textbf{Egocentric Gaze Estimation in the Wild.}
Out-of-distribution generalization on EgoExo4D is challenging under strictly online evaluation because the model must predict gaze accurately as the test distribution shifts across participants, sites, and tasks. Table~\ref{tab:egoexo4d_ood_generalization} shows that simple priors provide useful reference points but are inherently limited by static spatial biases. In contrast, our autoregressive model performs best on every split, reaching 70.58 F1 on IID while remaining strong under OOD participant, site, and task shifts with 68.22, 69.56, and 71.10 F1, respectively. This consistency, especially on OOD task, suggests that autoregressive modeling captures transferable attention dynamics rather than overfitting to domain-specific cues. The stable precision–recall profile further supports that the gains come from improved temporal reasoning and context integration, which is critical for real-world deployment where both environments and behaviors change unpredictably.

\begin{figure*}[t]
    \centering
    \includegraphics[width=0.9\linewidth]{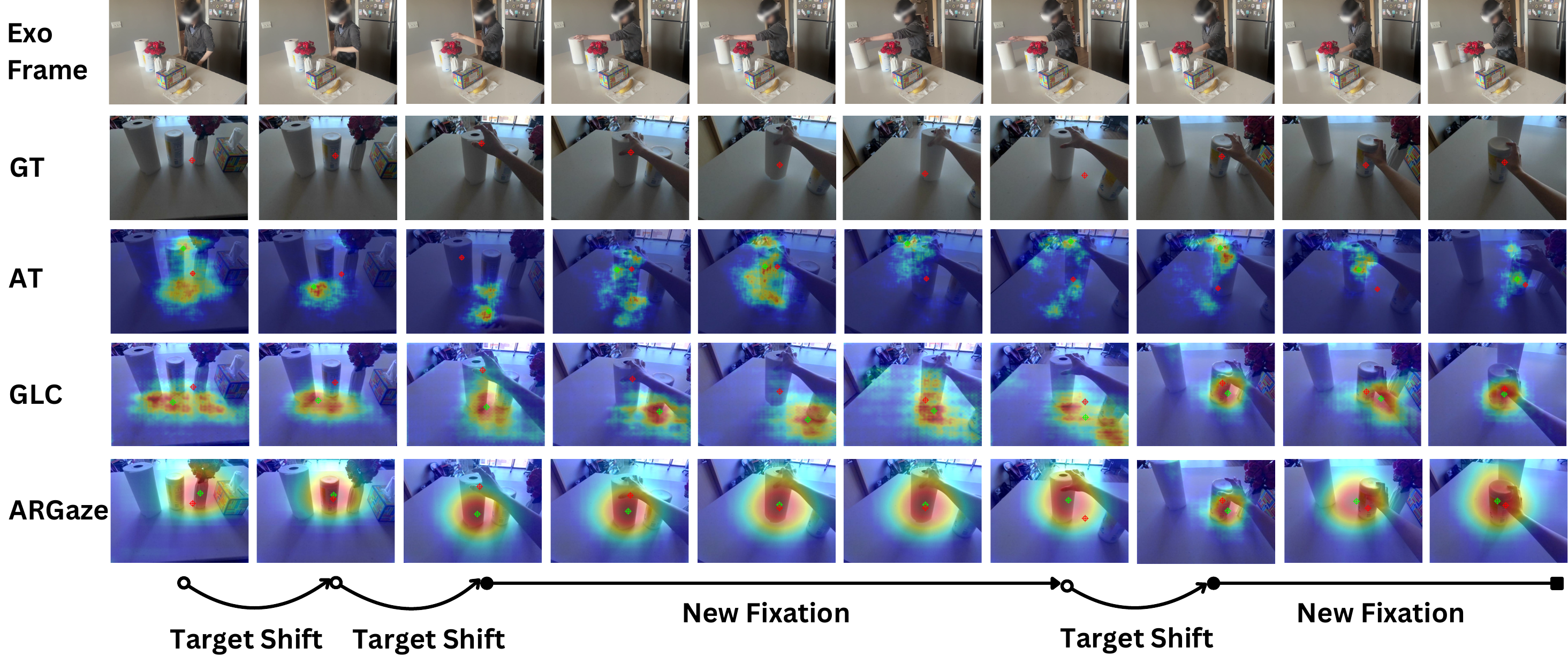}
    \vspace{-10pt}
    \caption{\textbf{Comparison of gaze heatmaps across methods during target shifts and new fixations.} ARGaze fuses visual cues with the historical continuity of the gaze trajectory, enabling rapid adaptation during target shift and stable fixation on task-relevant objects.}
    \label{fig:organize_table}
    \vspace{-15pt}
\end{figure*} 

\begin{table}[t]
    \centering
    \small
    \setlength{\tabcolsep}{6pt}
    \renewcommand{\arraystretch}{0.9}
    \caption{\textbf{History length ablation on EGTEA Gaze+.} $K$ denotes the number of past predicted heatmaps.}
    \begin{tabular}{lccc}
        \toprule
        \textbf{Variant} & \textbf{F1} $\uparrow$ & \textbf{Prec.} $\uparrow$ & \textbf{Rec.} $\uparrow$ \\
        \midrule
        Per-frame baseline, $K{=}0$ & 41.94 & 31.91 & 61.17 \\
        NoHist, $K{=}0$ (with template) & 42.56 & 32.27 & 62.49 \\
        History $K{=}1$ & 43.40 & 32.51 & \textbf{65.28} \\
        History $K{=}2$ & 43.55 & 34.26 & 59.76 \\
        History $K{=}3$ & \textbf{44.48} & \textbf{35.22} & 60.37 \\
        History $K{=}4$ & 43.62 & 34.46 & 59.40 \\
        History $K{=}5$ & 43.91 & 34.84 & 59.36 \\
        \bottomrule
    \end{tabular}
    \label{tab:ablation_history_k}
    \vspace{-19pt}
\end{table}
\begin{table}[t]
    \centering
    \small
    \setlength{\tabcolsep}{6pt}
    \renewcommand{\arraystretch}{0.9}
    \caption{\textbf{Template mechanism ablation on EGTEA Gaze+.} All performance metrics are reported in percentages (\%). We vary the template crop scale and related design choices while maintaining the autoregressive order at $K=3$.}
    \begin{tabular}{lccc}
        \toprule
        \textbf{Variant} & \textbf{F1} $\uparrow$ & \textbf{Prec.} $\uparrow$ & \textbf{Rec.} $\uparrow$ \\
        \midrule
        CropScale 0.0& 43.47 & 34.46 & 58.86 \\
        CropScale 0.25 & 44.05 & 34.81 & 59.96 \\
        CropScale 0.35 & \textbf{44.48} & \textbf{35.22} & 60.37 \\
        CropScale 0.55 & 43.90 & 34.62 & 59.98 \\
        CropScale 1.0 & 44.35 & 35.09 & 60.25 \\
        \midrule
        Multi-Scale [0.25, 0.35] & 43.46 & 34.07 & 60.00 \\
        Multi-Scale [0.15, 0.25, 0.35] & 43.47 & 34.19 & 59.67 \\
        \midrule
        CropScale 0.35, 2-Frames & 43.20 & 33.51 & 60.76 \\
        CropScale 0.35, 3-Frames & 43.53 & 33.86 & \textbf{60.94} \\
        \midrule
        ROI Embedding (0.35) & 43.83 & 34.42 & 60.32 \\
        CropEmd & 43.41 & 34.64 & 58.14 \\
        \bottomrule
    \end{tabular}
    \label{tab:ablation_template_cropscale}
    \vspace{-9mm}
\end{table}
\subsection{Analysis of the Autoregressive Model}
\textbf{Order of autoregression.}
We evaluate the efficacy of the autoregressive order $K$ by varying the number of historical gaze heatmaps provided to the decoder. As indicated in Table~\ref{tab:ablation_history_k}, the inclusion of temporal context leads to a significant enhancement in performance across all metrics compared to the per-frame baseline. Specifically, the framework reaches its performance peak at $K=3$, achieving an F1 score of 44.48\% and a precision of 35.22\%. This constitutes a substantial improvement over the 41.94\% F1 score and 31.91\% precision of the baseline model. While even a single frame of history yields an improvement, the performance reaches its optimal balance at three frames. Beyond this, as observed in the results for $K=4$ and $K=5$, the F1 score begins to fluctuate and experience a minor decline. This trend suggests that a history depth of three frames is sufficient for the model to capture necessary motion priors and maintain gaze continuity while avoiding the cumulative noise that often accompanies excessively long temporal windows.

\textbf{Template mechanism.}
The template mechanism serves as a localized visual prior by providing the decoder with cropped regions centered on previously predicted gaze coordinates. 
Our ablation results in Table~\ref{tab:ablation_template_cropscale} show that the crop scale affects performance.
Excluding the template tokens entirely leads to an F1 score drop to 43.47\%, which confirms that local spatial guidance is indispensable for accurate gaze estimation. We observe that a crop scale of 0.35 yields the optimal performance, achieving an F1 score of 44.48\% and a precision of 35.22\%. While a smaller scale of 0.25 captures fine-grained details, it may lack sufficient context, whereas larger scales such as 0.55 or 1.0 tend to introduce distracting background information that degrades the refinement process. Furthermore, although we explored multi-scale strategies to incorporate broader spatial information, the results indicate that a single-scale template at 0.35 is both more efficient and effective. This suggests that a well-calibrated local window provides enough visual evidence for the model to maintain focus on the relevant regions without the need for additional architectural complexity.


\begin{table}[t]
    \centering
    \footnotesize %
    \setlength{\tabcolsep}{0pt} 
    \caption{\textbf{Hand occlusion robustness on EGTEA Gaze+.} 
    We report results for subsets with and without visible hands.}
    \begin{tabular*}{\columnwidth}{@{\extracolsep{\fill}} lccc ccc }
        \toprule
        \multirow{2}{*}{\textbf{Method}} &
        \multicolumn{3}{c}{\textbf{With Hands}} &
        \multicolumn{3}{c}{\textbf{Without Hands}} \\
        \cmidrule(lr){2-4}\cmidrule(lr){5-7}
        & \textbf{F1}$\uparrow$ & \textbf{Prec.}$\uparrow$ & \textbf{Rec.}$\uparrow$ &
        \textbf{F1}$\uparrow$ & \textbf{Prec.}$\uparrow$ & \textbf{Rec.}$\uparrow$ \\
        \midrule
        AT            & 26.55 & 18.43 & 47.46 & 19.58 & 15.68 & 26.04 \\
        EgoM2P        & 41.72 & \textbf{37.06} & 48.59 & 30.46 & 27.77 & 34.39 \\  
        GLC     & 43.85 & 34.69 & 59.58 & 39.94 & 30.96 & 56.24 \\
        Ours (AR)     & \textbf{44.55} & {35.45} & \textbf{59.94} &
       \textbf{43.69} & \textbf{34.23} & \textbf{60.39} \\
        \bottomrule
    \end{tabular*}
    \label{tab:hand_occlusion_robustness}
    \vspace{-19pt}
    
\end{table}



\textbf{Robustness to Hand Dependency.}
In egocentric gaze estimation, hands act as a double-edged sword by providing essential behavioral cues while simultaneously introducing significant visual distractions. While models can exploit hand motion to infer intent, an over-reliance on these signals leads to two distinct failure modes. 
To analyze this, we partition the data into hand-visible and hand-invisible subsets using EgoHOS~\cite{zhang2022fine} with human verification.
As shown in Table~\ref{tab:hand_occlusion_robustness}, the baseline GLC model suffers from a performance collapse when hand cues are absent, with its F1 score dropping from 43.85 to 39.94. In contrast, our autoregressive model remains stable across both conditions because it leverages historical gaze trajectories to compensate for missing interaction evidence. 
This heavy dependency manifests conversely when hands are visible: they often introduce a spurious hand bias that leads models to erroneously track hand movement instead of the task-relevant object. 
Figure~\ref{fig:reading} qualitatively demonstrates this phenomenon, where GLC is distracted by manual manipulation while our structured dynamical model ensures a consistent fixation on the target; 
\sysname utilizes gaze history to filter visual distractions, achieving consistent performance across both hand-visible and hand-invisible settings.

\subsection{Real-World Testing}
To evaluate the practical utility of \sysname in authentic AR environments, we conducted streaming inference experiments using a Meta Quest 3 headset integrated with a Neon eye tracker, which provided high-fidelity ground truth at 30 FPS. We consider three daily activities: desk organizing, reading, and coffee making. 
Across all real-world streaming scenarios, \sysname consistently outperforms the GLC baseline, improving Heatmap F1 from 0.0411 to 0.4667. All efficiency numbers are measured on an NVIDIA RTX A6000 GPU: \sysname uses 32.3M parameters and runs at 89.56\,FPS with 11.17\,ms latency and 168.2\,MB peak memory.
Qualitatively, as shown in Figure~\ref{fig:reading} (reading) and Figure~\ref{fig:organize_table} (desk organizing), \sysname maintains stable fixations on task-relevant objects by effectively balancing instantaneous visual observation with the historical continuity of the user’s gaze target trajectory. These results further highlight the robustness and real-world generalization of ARGaze.
\section{Conclusion}
We present \sysname, a framework that reformulates egocentric gaze estimation as a strictly causal autoregressive sequence prediction problem. By conditioning on gaze target history, \sysname improves accuracy, robustness, and generalization across three benchmarks, including unseen tasks and novel scenes, while reducing reliance on hand–object cues and maintaining strong performance in hands-free settings. Its bounded history window makes \sysname efficient and practical for future real-time AR and assistive robotics.

\section*{Impact Statement}

This paper presents work aimed at advancing machine learning methods for online egocentric gaze estimation, with potential applications in augmented reality, assistive technologies, and human-centered AI systems. Improved real-time gaze prediction may enable more responsive and accessible interactive experiences.
At the same time, gaze estimation raises privacy and ethical concerns, as gaze behavior can reveal sensitive information about user intent and interests. Our work focuses on algorithmic and systems-level contributions and does not introduce new datasets involving personal identity or sensitive attributes. We encourage future research and deployment to prioritize privacy, user consent, and responsible use, ensuring that gaze-based technologies benefit users while minimizing potential harm.

{
    \small
    \bibliographystyle{icml2026}
    \bibliography{main}

\begin{thebibliography}{36}
\providecommand{\natexlab}[1]{#1}
\providecommand{\url}[1]{\texttt{#1}}
\expandafter\ifx\csname urlstyle\endcsname\relax
  \providecommand{\doi}[1]{doi: #1}\else
  \providecommand{\doi}{doi: \begingroup \urlstyle{rm}\Url}\fi

\bibitem[Alayrac et~al.(2022)Alayrac, Donahue, Luc, Miech, Barr, Hasson, Lenc,
  Mensch, Millican, Reynolds, et~al.]{alayrac2022flamingo}
Alayrac, J.-B., Donahue, J., Luc, P., Miech, A., Barr, I., Hasson, Y., Lenc,
  K., Mensch, A., Millican, K., Reynolds, M., et~al.
\newblock Flamingo: a visual language model for few-shot learning.
\newblock \emph{Advances in neural information processing systems},
  35:\penalty0 23716--23736, 2022.

\bibitem[Ba et~al.(2016)Ba, Kiros, and Hinton]{ba2016layer}
Ba, J.~L., Kiros, J.~R., and Hinton, G.~E.
\newblock Layer normalization.
\newblock \emph{arXiv preprint arXiv:1607.06450}, 2016.

\bibitem[Belardinelli(2024)]{belardinelli2024gaze}
Belardinelli, A.
\newblock Gaze-based intention estimation: principles, methodologies, and
  applications in hri.
\newblock \emph{ACM Transactions on Human-Robot Interaction}, 13\penalty0
  (3):\penalty0 1--30, 2024.

\bibitem[Beltagy et~al.(2020)Beltagy, Peters, and Cohan]{beltagy2020longformer}
Beltagy, I., Peters, M.~E., and Cohan, A.
\newblock Longformer: The long-document transformer.
\newblock \emph{arXiv preprint arXiv:2004.05150}, 2020.

\bibitem[Carreira \& Zisserman(2017)Carreira and Zisserman]{i3d}
Carreira, J. and Zisserman, A.
\newblock Quo vadis, action recognition? a new model and the i3d architecture.
\newblock In \emph{Proceedings of the IEEE Conference on Computer Vision and
  Pattern Recognition (CVPR)}, pp.\  6299--6308, 2017.

\bibitem[Damen et~al.(2022)Damen, Doughty, Farinella, Furnari, Kazakos, Ma,
  Moltisanti, Munro, Perrett, Price, et~al.]{damen2022rescaling}
Damen, D., Doughty, H., Farinella, G.~M., Furnari, A., Kazakos, E., Ma, J.,
  Moltisanti, D., Munro, J., Perrett, T., Price, W., et~al.
\newblock Rescaling egocentric vision: Collection, pipeline and challenges for
  epic-kitchens-100.
\newblock \emph{International Journal of Computer Vision}, 130\penalty0
  (1):\penalty0 33--55, 2022.

\bibitem[Deng et~al.(2024)Deng, Pan, Diao, Luo, Cui, Lu, Shan, Qi, and
  Wang]{deng2024nova}
Deng, H., Pan, T., Diao, H., Luo, Z., Cui, Y., Lu, H., Shan, S., Qi, Y., and
  Wang, X.
\newblock Autoregressive video generation without vector quantization.
\newblock \emph{arXiv preprint arXiv:2412.14169}, 2024.

\bibitem[Fathi et~al.(2012)Fathi, Li, and Rehg]{fathi2012learning}
Fathi, A., Li, Y., and Rehg, J.~M.
\newblock Learning to recognize daily actions using gaze.
\newblock In \emph{European Conference on Computer Vision}, pp.\  314--327.
  Springer, 2012.

\bibitem[Grauman et~al.(2022)Grauman, Westbury, Byrne, Chavis, Furnari,
  Girdhar, Hamburger, Jiang, Liu, Liu, et~al.]{grauman2022ego4d}
Grauman, K., Westbury, A., Byrne, E., Chavis, Z., Furnari, A., Girdhar, R.,
  Hamburger, J., Jiang, H., Liu, M., Liu, X., et~al.
\newblock Ego4d: Around the world in 3,000 hours of egocentric video.
\newblock In \emph{Proceedings of the IEEE/CVF conference on computer vision
  and pattern recognition}, pp.\  18995--19012, 2022.

\bibitem[Grauman et~al.(2024)Grauman, Westbury, Torresani, Kitani, Malik,
  Afouras, Ashutosh, Baiyya, Bansal, Boote, et~al.]{grauman2024egoexo4d}
Grauman, K., Westbury, A., Torresani, L., Kitani, K., Malik, J., Afouras, T.,
  Ashutosh, K., Baiyya, V., Bansal, S., Boote, B., et~al.
\newblock Ego-exo4d: Understanding skilled human activity from first-and
  third-person perspectives.
\newblock In \emph{Proceedings of the IEEE/CVF Conference on Computer Vision
  and Pattern Recognition}, pp.\  19383--19400, 2024.

\bibitem[Gupta et~al.(2024)Gupta, Tafasca, Farkhondeh, Vuillecard, and
  Odobez]{gupta2024mtgs}
Gupta, A., Tafasca, S., Farkhondeh, A., Vuillecard, P., and Odobez, J.-M.
\newblock Mtgs: A novel framework for multi-person temporal gaze following and
  social gaze prediction.
\newblock \emph{Advances in Neural Information Processing Systems},
  37:\penalty0 15646--15673, 2024.

\bibitem[Holmqvist et~al.(2011)Holmqvist, Nystr{\"o}m, Andersson, Dewhurst,
  Jarodzka, and Van~de Weijer]{holmqvist2011eye}
Holmqvist, K., Nystr{\"o}m, M., Andersson, R., Dewhurst, R., Jarodzka, H., and
  Van~de Weijer, J.
\newblock \emph{Eye tracking: A comprehensive guide to methods and measures}.
\newblock oup Oxford, 2011.

\bibitem[Huang et~al.(2018{\natexlab{a}})Huang, Cai, Li, and Sato]{huang2018}
Huang, Y., Cai, M., Li, Z., and Sato, Y.
\newblock Predicting gaze in egocentric video by learning task-dependent
  attention transition.
\newblock In \emph{Proceedings of the European Conference on Computer Vision
  (ECCV)}, pp.\  754--769, 2018{\natexlab{a}}.

\bibitem[Huang et~al.(2018{\natexlab{b}})Huang, Cai, Li, and
  Sato]{huang2018predicting}
Huang, Y., Cai, M., Li, Z., and Sato, Y.
\newblock Predicting gaze in egocentric video by learning task-dependent
  attention transition.
\newblock In \emph{Proceedings of the European conference on computer vision
  (ECCV)}, pp.\  754--769, 2018{\natexlab{b}}.

\bibitem[Lai et~al.(2024{\natexlab{a}})Lai, Liu, Ryan, and Rehg]{glc}
Lai, B., Liu, M., Ryan, F., and Rehg, J.~M.
\newblock In the eye of transformer: Global--local correlation for egocentric
  gaze estimation and beyond.
\newblock \emph{International Journal of Computer Vision}, 132\penalty0
  (3):\penalty0 854--871, 2024{\natexlab{a}}.

\bibitem[Lai et~al.(2024{\natexlab{b}})Lai, Ryan, Jia, Liu, and
  Rehg]{lai2024listen}
Lai, B., Ryan, F., Jia, W., Liu, M., and Rehg, J.~M.
\newblock Listen to look into the future: Audio-visual egocentric gaze
  anticipation.
\newblock In \emph{European Conference on Computer Vision (ECCV)}, pp.\
  192--210. Springer, 2024{\natexlab{b}}.
\newblock \doi{10.1007/978-3-031-72673-6_11}.

\bibitem[Lee et~al.(2024)Lee, Wang, Brown, Chu, Rodriguez, and
  Froehlich]{lee2024gazepointar}
Lee, J., Wang, J., Brown, E., Chu, L., Rodriguez, S.~S., and Froehlich, J.~E.
\newblock Gazepointar: A context-aware multimodal voice assistant for pronoun
  disambiguation in wearable augmented reality.
\newblock In \emph{Proceedings of the 2024 CHI Conference on Human Factors in
  Computing Systems}, pp.\  1--20, 2024.

\bibitem[Li et~al.(2025)Li, Chen, Wu, Zhao, Pollefeys, and Tang]{li2025egom2p}
Li, G., Chen, Y., Wu, Y., Zhao, K., Pollefeys, M., and Tang, S.
\newblock Egom2p: Egocentric multimodal multitask pretraining.
\newblock \emph{arXiv preprint arXiv:2506.07886}, 2025.

\bibitem[Li et~al.(2023{\natexlab{a}})Li, Li, Savarese, and Hoi]{li2023blip}
Li, J., Li, D., Savarese, S., and Hoi, S.
\newblock Blip-2: Bootstrapping language-image pre-training with frozen image
  encoders and large language models.
\newblock pp.\  19730--19742, 2023{\natexlab{a}}.

\bibitem[Li et~al.(2013)Li, Fathi, and Rehg]{li2013}
Li, Y., Fathi, A., and Rehg, J.~M.
\newblock Learning to predict gaze in egocentric video.
\newblock In \emph{Proceedings of the IEEE International Conference on Computer
  Vision (ICCV)}, pp.\  3216--3223, 2013.

\bibitem[Li et~al.(2018)Li, Liu, and Rehg]{li2018egtea}
Li, Y., Liu, M., and Rehg, J.~M.
\newblock In the eye of beholder: Joint learning of gaze and actions in first
  person video.
\newblock In \emph{Proceedings of the European conference on computer vision
  (ECCV)}, pp.\  619--635, 2018.

\bibitem[Li et~al.(2021)Li, Liu, and Rehg]{li2021eye}
Li, Y., Liu, M., and Rehg, J.~M.
\newblock In the eye of the beholder: Gaze and actions in first person video.
\newblock \emph{IEEE transactions on pattern analysis and machine
  intelligence}, 45\penalty0 (6):\penalty0 6731--6747, 2021.

\bibitem[Li et~al.(2023{\natexlab{b}})Li, Wang, Ma, Wang, and
  Meyer]{li2023swingaze}
Li, Y., Wang, X., Ma, Z., Wang, Y., and Meyer, M.~C.
\newblock Swingaze: Egocentric gaze estimation with video swin transformer.
\newblock In \emph{2023 IEEE 16th International Symposium on Embedded
  Multicore/Many-core Systems-on-Chip (MCSoC)}, pp.\  123--127,
  2023{\natexlab{b}}.
\newblock \doi{10.1109/MCSoC60832.2023.00026}.

\bibitem[Li et~al.(2023{\natexlab{c}})]{swingaze}
Li, Z. et~al.
\newblock Swingaze: Egocentric gaze estimation with video swin transformer.
\newblock \emph{arXiv preprint arXiv:2312.00000}, 2023{\natexlab{c}}.

\bibitem[Peng et~al.(2025)Peng, Hua, Liu, and Lu]{peng2025eye}
Peng, T., Hua, J., Liu, M., and Lu, F.
\newblock In the eye of mllm: Benchmarking egocentric video intent
  understanding with gaze-guided prompting.
\newblock \emph{arXiv preprint arXiv:2509.07447}, 2025.

\bibitem[Recasens et~al.(2015)Recasens, Khosla, Vondrick, and
  Torralba]{recasens2015they}
Recasens, A., Khosla, A., Vondrick, C., and Torralba, A.
\newblock Where are they looking?
\newblock \emph{Advances in neural information processing systems}, 28, 2015.

\bibitem[Ryan et~al.(2025)Ryan, Bati, Lee, Bolya, Hoffman, and
  Rehg]{ryan2025gaze}
Ryan, F., Bati, A., Lee, S., Bolya, D., Hoffman, J., and Rehg, J.~M.
\newblock Gaze-lle: Gaze target estimation via large-scale learned encoders.
\newblock In \emph{Proceedings of the Computer Vision and Pattern Recognition
  Conference}, pp.\  28874--28884, 2025.

\bibitem[Sim{\'e}oni et~al.(2025)Sim{\'e}oni, Vo, Seitzer, Baldassarre, Oquab,
  Jose, Khalidov, Szafraniec, Yi, Ramamonjisoa, et~al.]{simeoni2025dinov3}
Sim{\'e}oni, O., Vo, H.~V., Seitzer, M., Baldassarre, F., Oquab, M., Jose, C.,
  Khalidov, V., Szafraniec, M., Yi, S., Ramamonjisoa, M., et~al.
\newblock Dinov3.
\newblock \emph{arXiv preprint arXiv:2508.10104}, 2025.

\bibitem[Tran et~al.(2015)Tran, Bourdev, Fergus, Torresani, and
  Paluri]{tran20153dconv}
Tran, D., Bourdev, L., Fergus, R., Torresani, L., and Paluri, M.
\newblock Learning spatiotemporal features with 3d convolutional networks.
\newblock In \emph{Proceedings of the IEEE international conference on computer
  vision}, pp.\  4489--4497, 2015.

\bibitem[van~den Oord et~al.(2016{\natexlab{a}})van~den Oord, Kalchbrenner, and
  Kavukcuoglu]{pixelrnn}
van~den Oord, A., Kalchbrenner, N., and Kavukcuoglu, K.
\newblock Pixel recurrent neural networks.
\newblock \emph{arXiv preprint arXiv:1601.06759}, 2016{\natexlab{a}}.

\bibitem[van~den Oord et~al.(2016{\natexlab{b}})van~den Oord, Kalchbrenner,
  Vinyals, Espeholt, Graves, and Kavukcuoglu]{pixelcnn}
van~den Oord, A., Kalchbrenner, N., Vinyals, O., Espeholt, L., Graves, A., and
  Kavukcuoglu, K.
\newblock Conditional image generation with pixelcnn decoders.
\newblock \emph{arXiv preprint arXiv:1606.05328}, 2016{\natexlab{b}}.

\bibitem[Van Den~Oord et~al.(2017)Van Den~Oord, Vinyals, et~al.]{van2017vqvae}
Van Den~Oord, A., Vinyals, O., et~al.
\newblock Neural discrete representation learning.
\newblock \emph{Advances in neural information processing systems}, 30, 2017.

\bibitem[Vaswani et~al.(2017)Vaswani, Shazeer, Parmar, Uszkoreit, Jones, Gomez,
  Kaiser, and Polosukhin]{vaswani2017attention}
Vaswani, A., Shazeer, N., Parmar, N., Uszkoreit, J., Jones, L., Gomez, A.~N.,
  Kaiser, {\L}., and Polosukhin, I.
\newblock Attention is all you need.
\newblock \emph{Advances in neural information processing systems}, 30, 2017.

\bibitem[Yan et~al.(2021)Yan, Zhang, Abbeel, and Srinivas]{videogpt}
Yan, W., Zhang, Y., Abbeel, P., and Srinivas, A.
\newblock Videogpt: Video generation using vq-vae and transformers.
\newblock \emph{arXiv preprint arXiv:2104.10157}, 2021.

\bibitem[Zhang et~al.(2022)Zhang, Zhou, Stent, and Shi]{zhang2022fine}
Zhang, L., Zhou, S., Stent, S., and Shi, J.
\newblock Fine-grained egocentric hand-object segmentation: Dataset, model, and
  applications.
\newblock In \emph{European Conference on Computer Vision}, pp.\  127--145.
  Springer, 2022.

\bibitem[Zhang et~al.(2017)Zhang, Teck~Ma, Hwee~Lim, Zhao, and
  Feng]{zhang2017anticipation}
Zhang, M., Teck~Ma, K., Hwee~Lim, J., Zhao, Q., and Feng, J.
\newblock Deep future gaze: Gaze anticipation on egocentric videos using
  adversarial networks.
\newblock In \emph{Proceedings of the IEEE conference on computer vision and
  pattern recognition}, pp.\  4372--4381, 2017.

\end{thebibliography}
}

\newpage
\appendix
\onecolumn

\section{Evaluation Protocol and Split Construction}

\subsection{Hand Visible and Hand Absent Split Construction}
\label{app:hand_split}

We construct the hand visible and hand absent splits on EGTEA Gaze+ ~\cite{li2018egtea} using predicted hand masks from EgoHOS~\cite{zhang2022fine}. Our goal is to separate samples based on whether hands are present in the visual input, while ensuring that the split assignment is reliable enough to support a robustness analysis.

\begin{figure}[h]
    \centering
    \includegraphics[width=1.0\linewidth]{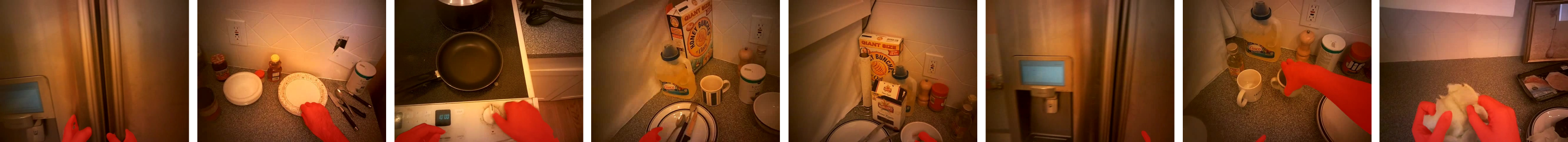}
    \caption{EgoHOS hand masks (red) overlaid on video frames from EGTEA.}
    \label{fig:egohos}
\end{figure}

We first validate the quality of EgoHOS on EGTEA using the Hand14K subset, which provides ground truth hand masks. We run EgoHOS inference on the Hand14K images and compare the predicted masks against the provided annotations under a binary hand segmentation protocol that merges all hand instances into a single foreground class. This evaluation yields an overall mean Intersection over Union of 0.8072, with 0.9937 Intersection over Union on background and 0.6207 on the hand class. Qualitatively, the predictions are stable and we do not observe systematic missed detections, which supports using EgoHOS to infer hand presence. Quantitative examples shows in Figure~\ref{fig:egohos} 

We then apply EgoHOS to EGTEA Gaze+ to obtain hand masks at the frame level. To further assess reliability under the target distribution, we randomly sample 1{,}000 frames from EGTEA Gaze+ and manually review the predicted masks. This review confirms that the predicted masks are consistent with the visual evidence, so we use them to define hand presence. Each frame is assigned to the hand visible split if EgoHOS predicts any hand pixels in that frame, and to the hand absent split otherwise. We compute all metrics separately on the resulting two frame level splits.

\subsection{EgoExo4D OOD Split Construction Details}
\label{app:egoexo_splits}

We construct a structured out of distribution benchmark on EgoExo4D to analyze robustness under controlled distribution shifts. Our goal is to isolate the effect of changing the collection site, the task category, or the participant identity, while keeping the remaining factors well represented in the training pool. All splits are created on the egocentric gaze subset, and we retain only clips with valid gaze supervision, including both personalized and general gaze annotations. After filtering, the resulting corpus contains 49{,}489 labeled samples spanning 43 fine grained tasks, 12 collection sites, and 377 unique participants. A coverage analysis shows that personalized and general gaze have very similar distributions across tasks, sites, and participants, so we treat both as equally valid supervision and discard only the small fraction of clips without any gaze.

To ensure that each split reflects a meaningful and non degenerate distribution shift, we use a split planner that selects hold out factors only when the remaining data can still provide adequate coverage. When holding out a collection site, we require that most tasks appearing at that site also appear in multiple other sites with sufficient samples, so that the model can learn the same task semantics during training without seeing the held out environment. When holding out a task, we require that the remaining pool contains diverse instances of that task across several sites and several participants with enough total samples, which encourages the training distribution to cover a broad range of contexts. When holding out a participant, we require that the participant performs largely overlapping tasks and appears in similar sites compared to the rest of the dataset, and we also require a minimum number of samples from that participant, which prevents trivial cases where the held out individual is associated with idiosyncratic activities that never occur elsewhere. In our benchmark, the split planner selects three held out tasks, namely \emph{Cooking Noodles}, \emph{Cooking Tomato \& Eggs}, and \emph{Covid-19 Rapid Antigen Test}, holds out two collection sites, namely \emph{georgiatech} and \emph{fair}, and chooses held out participants from frequent users that satisfy the overlap constraints.

Given the selected held out factors, we build disjoint evaluation splits that isolate each source of shift. The site shift split contains all samples recorded at the held out sites, so performance reflects generalization to unseen environments and scene statistics. The task shift split contains all samples from the held out tasks after removing any samples already assigned to the site shift split, so the evaluation reflects generalization to unseen task categories without confounding the result with unseen sites. The participant shift split contains all samples from the held out participants after removing samples assigned to the site shift and task shift splits, so it measures generalization to unseen individuals performing tasks and in environments that remain represented in training. From the remaining pool, we then construct an independent and identically distributed test split using stratified sampling over task and site so that its empirical distribution closely matches training. We also sample a validation set from the same remaining pool using the same stratified procedure. All remaining samples form the training set.

To prevent temporal leakage, we enforce a strict group by take constraint throughout the entire process, meaning that all frames and derived clips from a single recording session are assigned to the same split. After constructing the raw splits, we optionally subsample within each split using stratified sampling over task and site to reduce training cost while preserving relative frequencies. Unless otherwise stated, we train each model once on TRAIN, select hyperparameters on VAL, and evaluate the same checkpoint on TEST\_IID, TEST\_OOD\_SITE, TEST\_OOD\_TASK, and TEST\_OOD\_PARTICIPANT.

\section{Implementation Details.}

\subsection{ARGaze Implementation Details.}
\label{app:argaze_implementation}
Following prior work \cite{glc}, our model produces a $64 \times 64$ gaze heatmap. We utilize an input resolution of $224 \times 224$ for both the egocentric frames and the corresponding tracking-aware template crops, balancing visual detail with computational efficiency. Our experiments are conducted using frozen DINOv3 backbones~\cite{simeoni2025dinov3}, specifically the ViT-S/16 architecture. Using a patch size of $16 \times 16$, the input image is partitioned into a $14 \times 14$ grid of patches. We employ an internal dimension of $d_{model} = 384$, with 3 transformer decoder layers and 8 attention heads. The framework is trained using the AdamW optimizer with a base learning rate of $1e-4$ and a batch size of 16. Specifically, we train for 25 epochs on the EGTEA Gaze+ dataset and 15 epochs on Ego4D and EgoExo4D. To bridge the gap between training and inference, we implement scheduled sampling using a linear ramp that increases the sampling probability $p_{ss}$ from 0.0 to 0.3 over the first 10 epochs. We maintain a historical window of $k=3$ previous heatmaps and use a single template crop whose side length is set to 0.35 of the input image size to provide localized spatial priors.

\subsection{GLC Online Variant Implentation Details}
\label{app:glc_online}
To obtain an online-capable model suitable for fair comparision, we derive a strictly causal variant of GLC while keeping the overall architecture unchanged. Our modifications are limited to temporal computations so that the prediction at timestep $t$ depends only on inputs up to and including $t$.

We first make the temporal processing in the \texttt{PatchEmbed} and \texttt{GlobalEmbed} layers causal. Concretely, we convert each 3D convolution to a causal operator by replacing symmetric temporal padding with asymmetric padding. With this change, the receptive field of a convolution evaluated at time $t$ contains only frames whose indices are less than or equal to $t$, which removes any dependence on future frames introduced by padding.

We then enforce causality throughout the Transformer attention stack. For every Transformer block used in \texttt{MultiScaleBlock} and \texttt{GlobalLocalBlock}, we apply a causal attention mask in the scaled dot product attention. This mask prevents any token associated with timestep $t$ from attending to tokens from timesteps greater than $t$, while still allowing attention over all past and current timesteps. Together, these changes guarantee that the model output at time $t$ is a function only of past and present observations, enabling streaming deployment without temporal leakage.

\section{Additional Ablations and Analyses}
\subsection{Decoder Depth}
We study the effect of decoder depth by varying the number of decoder layers $N \in \{1,3,6\}$ while keeping all other settings fixed on EGTEA Gaze+. As shown in Table~\ref{tab:decoder_depth}, increasing depth from $N{=}1$ to $N{=}3$ improves overall performance, with F1 rising from $43.243\%$ to $44.484\%$ and precision from $32.945\%$ to $35.216\%$, although recall slightly drops from $62.908\%$ to $60.371\%$. Further increasing depth to $N{=}6$ does not yield additional gains, at the F1 score marginally decreases to $44.084\%$ and both precision ($34.866\%$) and recall ($59.929\%$) decline compared to $N{=}3$. These results suggest a moderate decoder depth ($N{=}3$) offers the best trade-off, improving precision and F1 without incurring the diminishing returns observed at greater depth.


\subsection{Scheduled sampling}
Scheduled sampling (SS) is introduced to mitigate exposure bias by gradually replacing ground-truth inputs with model predictions during training. As shown in Table~\ref{tab:ss_ablation_egogaze}, enabling SS consistently improves performance over the no-SS baseline. In particular, F1 increases from 42.30\% at SS=0.0 to a peak of 44.48\% at SS=0.3, indicating that a moderate amount of sampling helps the model better handle test-time autoregressive conditions. Increasing SS further yields diminishing returns, as SS=0.5 achieves a comparable F1 (44.18\%) while providing the highest recall (63.11\%), whereas larger ratios slightly favor precision but reduce recall, resulting in near-identical F1. Overall, these results suggest that a moderate SS ratio=0.3 offers the best trade-off, improving robustness without overly amplifying prediction noise.

\begin{table}[t]
\centering
\small
\setlength{\tabcolsep}{3pt}
\renewcommand{\arraystretch}{1.10}

\begin{minipage}[t]{0.30\linewidth}
\centering
\caption{\textbf{Embedding ablation} on EGTEA Gaze+ (\%).}
\label{tab:ablation_embedding_design}
\begin{tabular}{lccc}
\toprule
\textbf{Variant} & \textbf{F1} & \textbf{P} & \textbf{R} \\
\midrule
No Embeddings & 43.23 & 34.00 & 59.35 \\
No Temporal & 44.19 & 34.64 & \textbf{61.03} \\
No TokenType & 43.70 & 33.98 & 61.23 \\
TT + Temp & \textbf{44.48} & \textbf{35.22} & 60.37 \\
\bottomrule
\end{tabular}
\end{minipage}\hspace{0.02\linewidth}%
\begin{minipage}[t]{0.30\linewidth}
\centering
\caption{\textbf{Scheduled sampling} ablation on EGTEA Gaze+ (\%).}
\label{tab:ss_ablation_egogaze}
\begin{tabular}{lccc}
\toprule
\textbf{SS} & \textbf{F1} & \textbf{P} & \textbf{R} \\
\midrule
0.0 & 42.30 & 32.84 & 59.40 \\
0.3 & \textbf{44.48} & \textbf{35.22} & 60.37 \\
0.5 & 44.18 & 33.98 & \textbf{63.11} \\
0.7 & 44.00 & 34.66 & 60.22 \\
1.0 & 43.99 & 35.04 & 59.11 \\
\bottomrule
\end{tabular}
\end{minipage}\hspace{0.02\linewidth}%
\begin{minipage}[t]{0.30\linewidth}
\centering
\caption{\textbf{Decoder depth} $N$ on EGTEA Gaze+.}
\label{tab:decoder_depth}
\begin{tabular}{lccc}
\toprule
\textbf{$N$} & \textbf{F1} & \textbf{P} & \textbf{R} \\
\midrule
1 & 43.24 & 32.95 & 62.91 \\
3 & \textbf{44.48} & \textbf{35.22} & 60.37 \\
6 & 44.08 & 34.87 & 59.93 \\
\bottomrule
\end{tabular}
\end{minipage}

\end{table}


\subsection{Embedding Design} We investigate the impact of embedding designs on the framework's ability to disambiguate information within the autoregressive sequence. As illustrated in Table~\ref{tab:ablation_embedding_design}, removing both temporal and token-type embeddings leads to a noticeable decline in the F1 score to 43.23\%, which underscores the necessity of providing structural and categorical cues to the decoder. Removing temporal embeddings alone results in a drop in F1 and precision.
This suggests that while the model retains its basic predictive capacity, it loses the refined chronological order essential for precise gaze forecasting. Similarly, excluding token-type embeddings hampers the model's ability to distinguish between current visual features and historical template priors, leading to suboptimal performance. The combined use of both temporal and token-type embeddings achieves the superior F1 score of 44.48\% and precision of 35.22\%. These results confirm that explicitly encoding the identity and timing of tokens allows the model to more effectively leverage the complex dependencies within the autoregressive window for robust gaze prediction.


\section{Additional Results}
\subsection{EgoExo4D Per Task Results}
Table~\ref{tab:egoexo4d_gaze_results} reports per-task performance under IID and two OOD settings (site shift and task shift). Overall, the model achieves consistently strong results across common activities, with F1 scores mostly in the mid-to-high 60s/70s, indicating reliable discrimination of gaze targets across diverse scenarios. Notably, OOD generalization under \textit{site shift} is competitive with IID performance for major tasks (e.g., Cooking and Bike Repair), suggesting that changes in capture environment have a limited impact on accuracy. The largest performance drops occur in long-tail tasks with substantially fewer evaluation clips (e.g., Soccer, Dance, Rock Climbing), where both F1 decreases and L2 error increases, pointing to data scarcity and task-specific variability as the primary bottlenecks rather than domain shift itself. Finally, the \textit{OOD (Task)} setting on Cooking remains close to IID, further supporting the robustness of the proposed approach in practical cross-domain deployments.

\begin{table}[ht]
\centering
\caption{Performance comparison across different tasks and settings (IID vs. OOD)}
\label{tab:egoexo4d_gaze_results}
\begin{small}
\begin{tabular}{@{}llccccc@{}}
\toprule
\textbf{Model/Setting} & \textbf{Task} & \textbf{F1} $\uparrow$ & \textbf{Recall} $\uparrow$ & \textbf{Precision} $\uparrow$ & \textbf{L2} $\downarrow$ & \textbf{Clips} \\ \midrule
\textit{OOD (Site)} & Cooking & 72.72 & 68.14 & 78.03 & 7.39 & 30,969 \\
 & Bike Repair & 68.17 & 64.48 & 72.39 & 8.89 & 15,408 \\
 & Health & 64.71 & 60.01 & 70.38 & 9.25 & 14,229 \\  \midrule
 \textit{OOD (Task)} & Cooking & 71.16 & 66.59 & 76.67 & 7.82 & 60,516 \\
 \midrule
\textit{IID} & Cooking & 71.81 & 66.98 & 77.51 & 7.78 & 37,188 \\
 & Music & 67.24 & 62.99 & 72.16 & 8.97 & 8,127 \\
 & Health & 70.59 & 65.78 & 76.33 & 7.72 & 6,759 \\
 & Bike Repair & 71.22 & 65.71 & 77.81 & 7.68 & 3,960 \\
 & Basketball & 71.09 & 66.77 & 76.04 & 8.09 & 3,762 \\
 & Soccer & 57.14 & 54.49 & 60.17 & 12.00 & 324 \\
 & Dance & 54.75 & 51.97 & 57.88 & 13.29 & 207 \\
 & Rock Climbing & 64.36 & 61.58 & 67.47 & 10.14 & 162 \\ 
 
\bottomrule
\end{tabular}
\end{small}
\end{table}

\section{Qualitative Results}
\begin{figure}[t]
    \centering
    \includegraphics[width=0.8\linewidth]{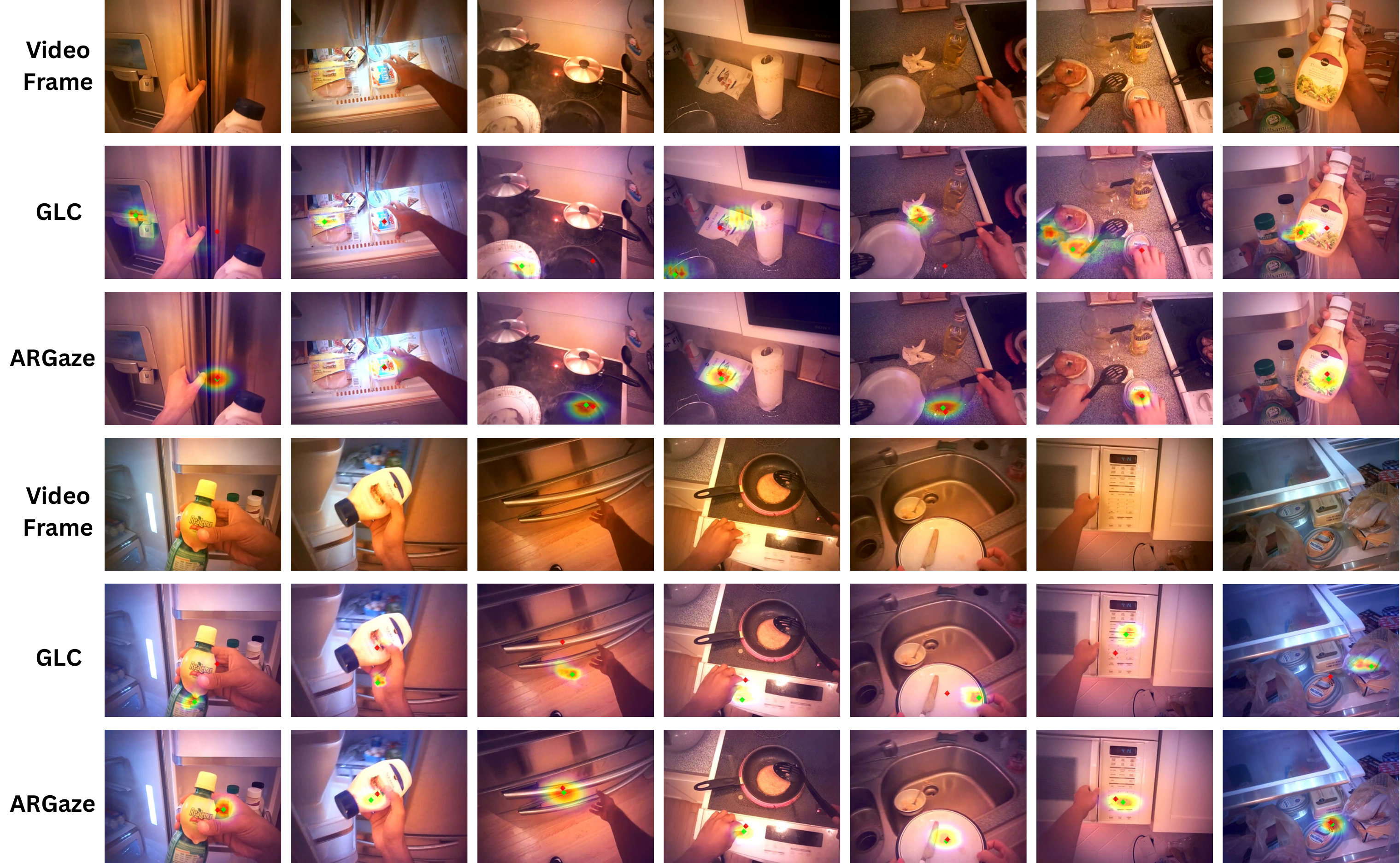}
    \caption{Visualization of baselines and \sysname on Egtea dataset.}
    \label{fig:visual}
\end{figure}

\begin{figure}[t]
    \centering
    \includegraphics[width=0.8\linewidth]{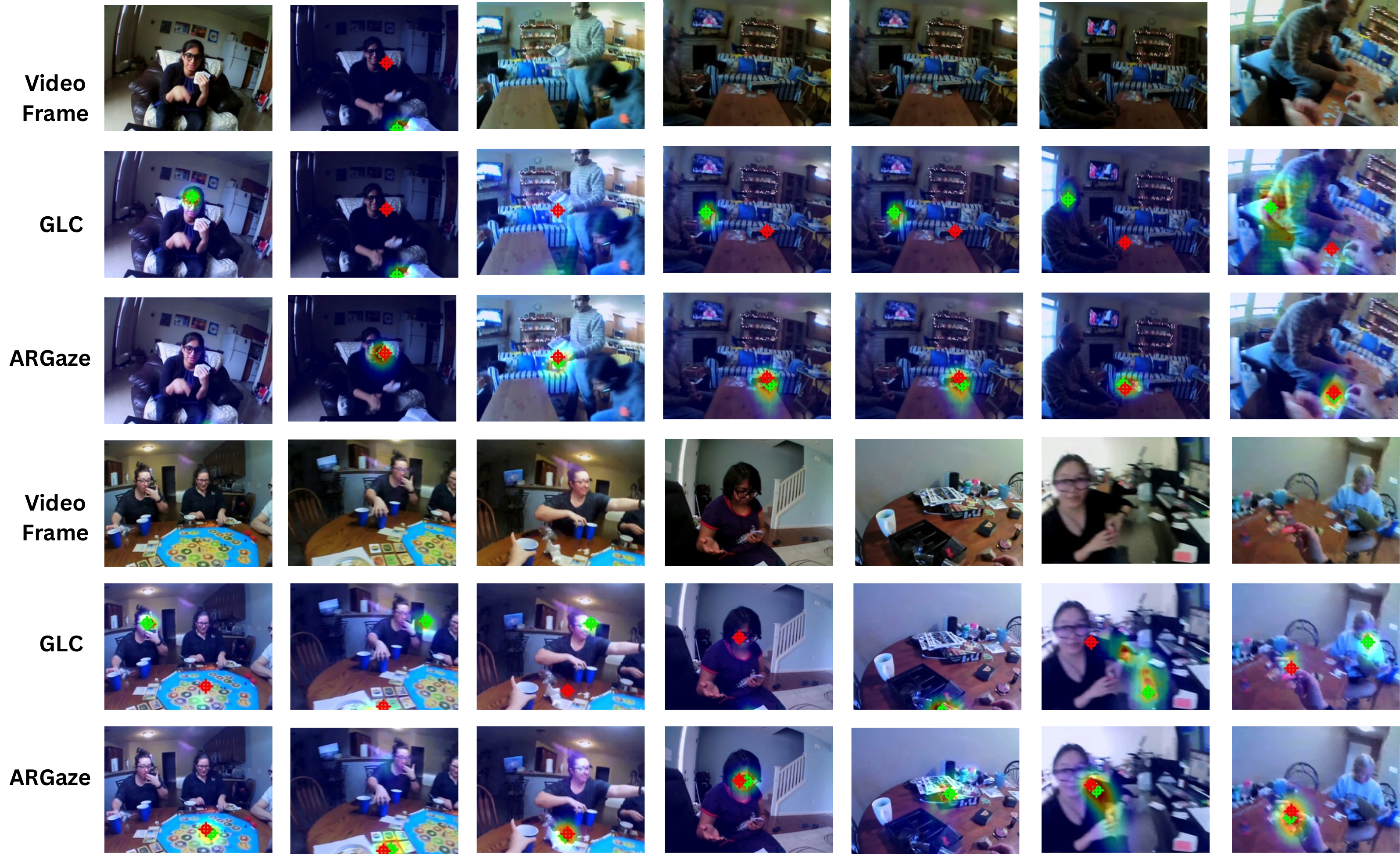}
    \caption{Visualization of baselines and \sysname on Ego4D dataset.}
    \label{fig:ego4d_visual}
\end{figure}

Figure~\ref{fig:visual} and Figure~\ref{fig:ego4d_visual} visualizes predicted gaze heatmaps for the baseline (GLC) and \sysname under diverse real-world scenarios. Compared to GLC, ARGaze produces more compact and task-aligned activations that consistently concentrate on the interacted objects (e.g., handles, utensils, containers) rather than drifting to hands or nearby clutter. Across challenging moments with rapid viewpoint changes and partial occlusions, ARGaze maintains stronger temporal consistency, yielding stable fixations that better reflect goal-directed attention in egocentric activities.

\section{Failure Case Analysis}
\begin{figure}[t]
    \centering
    \includegraphics[width=0.8\linewidth]{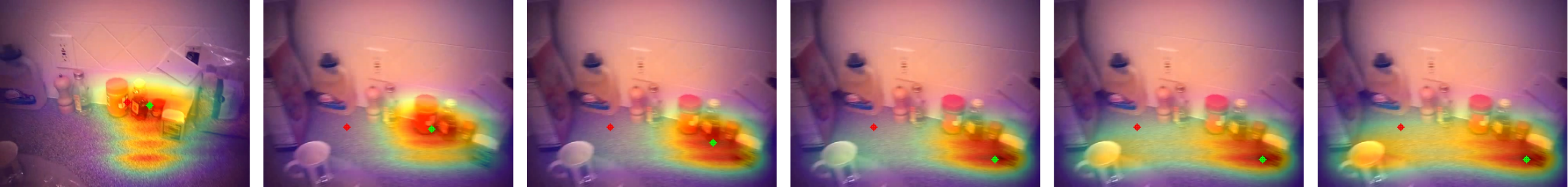}
    \caption{Failure case: rapid head motion. Under abrupt viewpoint changes and motion blur, the predicted gaze heatmap can temporarily drift before re-locking onto the correct target, leading to short-term mis-localization.}
    \label{fig:failure_1}
\end{figure}

\begin{figure}
    \centering
    \includegraphics[width=0.8\linewidth]{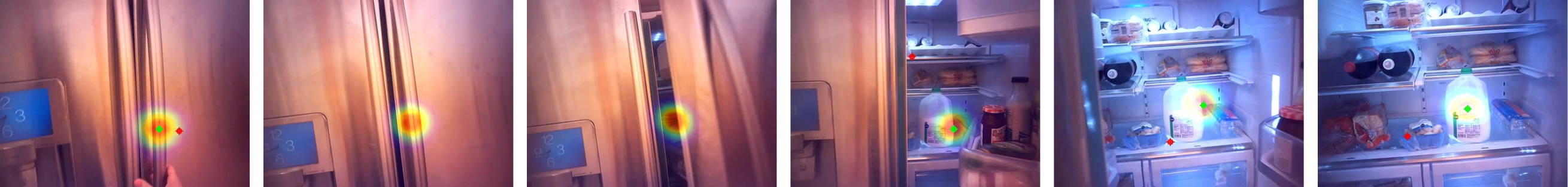}
    \caption{Failure case: saliency bias under weak cues. When task-relevant cues are ambiguous (e.g., no active interaction or distinctive appearance), the model may bias toward salient objects, producing attention on high-contrast distractors instead of the real target.}
    \label{fig:failure_2}
\end{figure}

We observe two representative failure modes in real-world streaming. First, under rapid head motion or abrupt viewpoint changes, the model may exhibit short-term drift before re-locking onto the correct target, likely due to motion blur and reduced temporal correspondences across frames (see Figure~\ref{fig:failure_1}). Second, when the scene provides weak task cues (e.g., the target is not actively interacted with or lacks distinctive appearance), the prediction can bias toward visually salient objects, leading to attention being attracted to high-contrast regions rather than the intended gaze target (see Figure~\ref{fig:failure_2}). These cases suggest that incorporating stronger motion compensation and additional task/context signals could further improve robustness.

\newpage
\section{Differences with Egocentric Gaze Anticipation}
\label{subsec_anticipation}

While our work focuses on egocentric gaze estimation, it is important to distinguish this from the related but fundamentally different task of gaze anticipation. Gaze estimation localizes fixation at the current timestamp $t$ given observations up to $t$, requiring accurate real-time localization as frames arrive. In contrast, gaze anticipation forecasts gaze location at a future timestamp $t+\delta$ based on past history, shifting the problem from localization to intent prediction. This temporal displacement introduces qualitatively different challenges: anticipation must reason about future actions and environmental changes, whereas estimation focuses on interpreting the current visual scene.

Early anticipation works, such as \textit{Deep Future Gaze} \cite{zhang2017anticipation}, employed Generative Adversarial Networks (GANs) to synthesize future frames and subsequently predict gaze on the hallucinated content. More recently, Lai et al. \cite{lai2024listen} highlighted the limitations of vision-only approaches in forecasting, introducing the Contrastive Spatial-Temporal Separable framework to leverage audio-visual cues for identifying out-of-view triggers (e.g., a sudden sound). Although anticipation models are inherently causal, their focus on long-term forecasting and multi-step trajectories makes them fundamentally different in both objective and evaluation metrics. Consequently, while anticipation provides valuable insights into temporal gaze modeling, direct performance comparison is inappropriate due to divergent task definitions and deployment requirements.

\end{document}